\documentclass[10pt,journal, comsoc]{IEEEtran} % 10pt, conference
\usepackage{scalerel,xparse}

\usepackage[draft]{todonotes}   % notes showed

\usepackage{color}
\usepackage{cite}
\usepackage{enumitem}

\ifCLASSINFOpdf
  \graphicspath{{../img/}}
\else
\fi
\usepackage{amssymb} %to have \mathbb{R}
\usepackage{amsmath}
\usepackage{newtxtext,newtxmath}
\ifCLASSOPTIONcompsoc
  \usepackage[caption=false,font=normalsize,labelfont=sf,textfont=sf]{subfig}
\else
  \usepackage[caption=false,font=footnotesize]{subfig}
\fi
\usepackage{booktabs}
\usepackage{multirow}
\usepackage{soul}
\usepackage{lipsum}
\usepackage{tikz}

\begin{document}

%\title{COTORRA:\emojicotorra{}{}\\COntext-aware Testbed fOR Robotic Applications }
\title{COTORRA:\\COntext-aware Testbed fOR Robotic Applications }
%\title{COTORRA:\emojicotorra{}{}\\COmodity Testbed fOR Robotic Applications }

\author{Milan~Groshev,~\IEEEmembership{Member,~IEEE,}
        Jorge~Martín-Pérez,~\IEEEmembership{Member,~IEEE,}
        Kiril~Antevski,~\IEEEmembership{Member,~IEEE,}% <-this % stops a space
}

\author{
    \IEEEauthorblockN{
        Milan~Groshev,
        Jorge~Martín-Pérez,
        Kiril Antevski, 
        Antonio de la Oliva, 
        Carlos J. Bernardos, 
        }

\thanks{M. Groshev, J. Mart\'in-P\'erez, K. Antevski, A. de la Oliva and Carlos J. Bernardos are with the Department
of Telematics Engineering at Universidad Carlos III de Madrid, Spain. e-mail: mgroshev@pa.uc3m.es, jmartinp@it.uc3m.es, kantevsk@pa.uc3m.es, aoliva@it.uc3m.es, cjbc@it.uc3m.es}% <-this % stops a space
\thanks{Work partially funded by the EU H2020
5GROWTH Project (grant no. 856709) and by the H2020
collaborative Europe/Taiwan research project 5G-DIVE
(grant no. 859881).
}% <-this % stops a space
%\thanks{Manuscript received April 19, 2005; revised August 26, 2015.}
}

% \author{
%     \IEEEauthorblockN{
%         Milan~Groshev\IEEEauthorrefmark{1},
%         Jorge~Martín-Pérez\IEEEauthorrefmark{1},
%         Kiril Antevski\IEEEauthorrefmark{1}, 
%         Antonio de la Oliva\IEEEauthorrefmark{1}, 
%         Carlos J. Bernardos\IEEEauthorrefmark{1}, 
%         }
%     \IEEEauthorblockA{
%         \IEEEauthorrefmark{1}University Carlos III of Madrid, Spain
%     }
% }

\markboth{Journal of \LaTeX\ Class Files,~Vol.~14, No.~8, August~2015}%
{Shell \MakeLowercase{\textit{et al.}}: Bare Demo of IEEEtran.cls for IEEE Journals}

\maketitle

%% -- DO NOT REMOVE ME --
%% 75-100 words
\begin{abstract}
%Edge and Fog computing can efficiently improve robotic systems, but such improvements are conditional on an effective testbed at the edge of the network. 
Edge \& Fog computing have received considerable attention as promising candidates for the evolution of robotic systems. In this letter, we propose COTORRA, an Edge \& Fog driven robotic testbed that combines context information with robot sensor data to validate innovative concepts for robotic systems prior to being applied in a production environment.
In lab/university, we established COTORRA as an easy applicable and modular testbed on top of heterogeneous network infrastructure. 
COTORRA is open for pluggable robotic applications.
To verify its feasibility and assess its performance, we ran set of experiments that show how autonomous navigation applications can achieve target latencies bellow 15ms or perform an inter-domain (DLT) federation within 19 seconds.
% with the use of Distributed Ledger Technologies.
% Through experimentation we demonstrate how the testbed can be used to validate the applicability of federation and placement mechanisms in robotic systems. We show how autonomous navigation applications can achieve target latencies bellow 15ms. Moreover, field tests confirm the testbed capabilities to validate and test the usage of Distributed Ledger Technology to perform inter-domain federation within 19 seconds. 
\end{abstract}


\IEEEpeerreviewmaketitle

\section{Introduction}
Networked robotics combines robotics with communication networks to enhance capabilities of robots. 
A clear example is Cloud robotics~\cite{cloudRobo}, which aims to integrate Cloud computing resources in robotics systems to increase re-configurability as well as to decrease the complexity and cost of robots. Thanks to Edge~\cite{edge.computing}
\& Fog~\cite{fog.computing} computing,
real-time applications as autonomous navigation will run closer to the robots,
and the robot-to-cloud communication will be enhanced thanks to improved
latency and reliability. Since the robot-to-cloud communication will traverse fewer network links, higher bandwidth rates, and bounded jitter are achievable. Moreover, real-time context information about the robots is expected to be available in the Edge, allowing dynamic adaptation of application's logic to the actual status of the communication (e.g., radio channel)~\cite{net.and.dev.5g}. Therefore, executing robotics applications in the Edge of the network is considered as a potential evolution of robotics system.

%In robotic systems, Edge~\cite{edge.computing} and Fog~\cite{fog.computing} computing can enable hard real-time applications to execute closer to the robots resulting in more predictable communication and overall better end-to-end system performance. Since the robot sensor data doesn't need to traverse many underlying networks, low-latency, high bandwidth and bounded jitter are achievable. Moreover, real-time context information about the robots is expected to be available in the Edge, allowing dynamic adaptation of application's logic to the actual status of the communication (e.g., radio channel)~\cite{net.and.dev.5g}. Therefore, executing robotics applications in the Edge of the network is consider as potential evolution of robotics system. 
Most of the existing work focuses on Cloud robotics testbeds or platforms~\cite{cloud.testbed,CLORO,vc.bots,openUAV,rapyuta,rospeex}. In~\cite{cloud.testbed}, the authors indicate the feasibility and challenges of a real world Cloud robotics implementation over Cloud testbed infrastructure. Besides,~\cite{CLORO} suggests a Cloud robotics testbed that can be effective for developing not only robot-related experiments but also network applications. By providing a Cloud robotics testbed for mobile robots, VC-bots~\cite{vc.bots} tries to address some important factors such as mobility, traffic scenario, protocol, cloud resources and localization awareness. Furthermore, due to the set of constrains (e.g., cost, time, safety) when performing physical testing of Unamanned Areal Vehicles,~\cite{openUAV} develops an Cloud-enabled, open-source simulation testbed for UAVs, thus reducing the entry barrier of UAV development and research. Finally,~\cite{rapyuta} and~\cite{rospeex} propose Cloud robotics platforms that address number of implementation-related issues.

However, it is relevant to note that all the existing robotics testbeds or platforms depend on the use of Cloud computing. Unfortunately, no Edge \& Fog robotics testbeds are explicitly mentioned in the existing literature, despite the need for an analysis on implementation issues, standards and validation for Edge robotics applications. There is a lack of practical experience on context-aware real-time environments in robotics systems. That leads to limited applicability in production systems and final products. 
% Motivated by this, 
% In this work we propose COTORRA, a real-time context-aware testbed architecture that we developed and validated through implementation of two different applications on top of it. 

In this work we propose COTORRA, a real-time context-aware testbed architecture that we implement and validate. This testbed is particularly suitable for time-sensitive robotic applications and mobile robots in Edge \& Fog environment. 
Additionally, to consider the time-sensitivity in computational offloading decisions, the COTORRA testbed offers network emulation capabilities.
To validate COTORRA, we implemented two applications
on top of it: an orchestration and a federation solution for robotic systems. The experiments and results
illustrate how COTORRA can be used to enhance robotic applications.

% %To validate robotic applications on top of COTORRA, we conduct experiments applying federation and placement mechanisms to robotic systems, and the results illustrate how COTORRA can be used to enhance robotic applications.

%Motivated by this, we propose COTORRA, a real-time context-aware testbed for robotic systems, designed to cope with the dynamic, heterogeneous and mobile Edge and Fog environment. This testbed is particularly suitable for time-sensitive robotic applications and mobile robots. 
%Additionally, to consider the time-sensitivity in computational offloading decisions, the testbed offers network emulation capabilities. To validate robotic applications on top of COTORRA, we conduct experiments applying federation and placement mechanisms to robotic systems, and the results illustrate how COTORRA can be used to enhance robotic applications.   

The remainder of this letter is organized as follows: we propose our real-time context-aware testbed for robotic applications in Sec.~\ref{sec:system}, and its possible implementation in Sec.~\ref{sec:implementation}. Finally, we evaluate the testbed in Sec.~\ref{sec:validation} and we conclude this letter in Sec.~\ref{sec:conclusion}.

\section{COTORRA system design}
\label{sec:system}

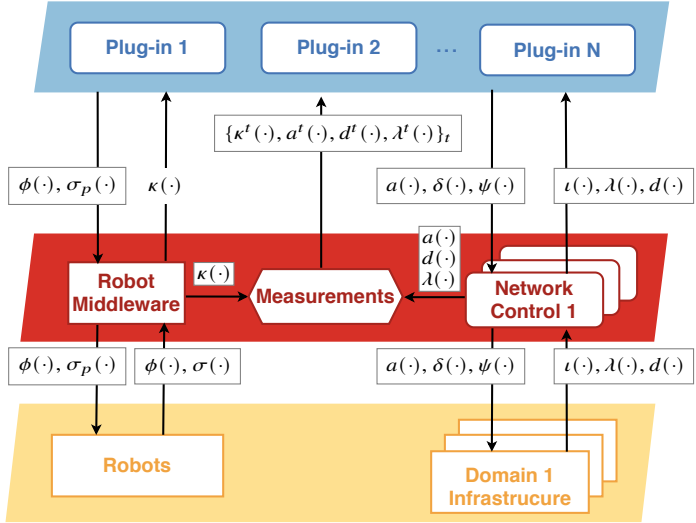
\begin{figure}[t]
    \begin{tikzpicture}[font=\scriptsize]
        % VARIABLES
        \pgfmathsetmacro{\below}{-1.4}
        \pgfmathsetmacro{\aboveAbove}{1.7}
        \pgfmathsetmacro{\aboveBelow}{1}
        \pgfmathsetmacro{\middle}{-.47}
        \pgfmathsetmacro{\middleAbove}{-.17}
        % PLOT
        \node[inner sep=0pt] (arch) at (0,0) {\includegraphics[width=\columnwidth]{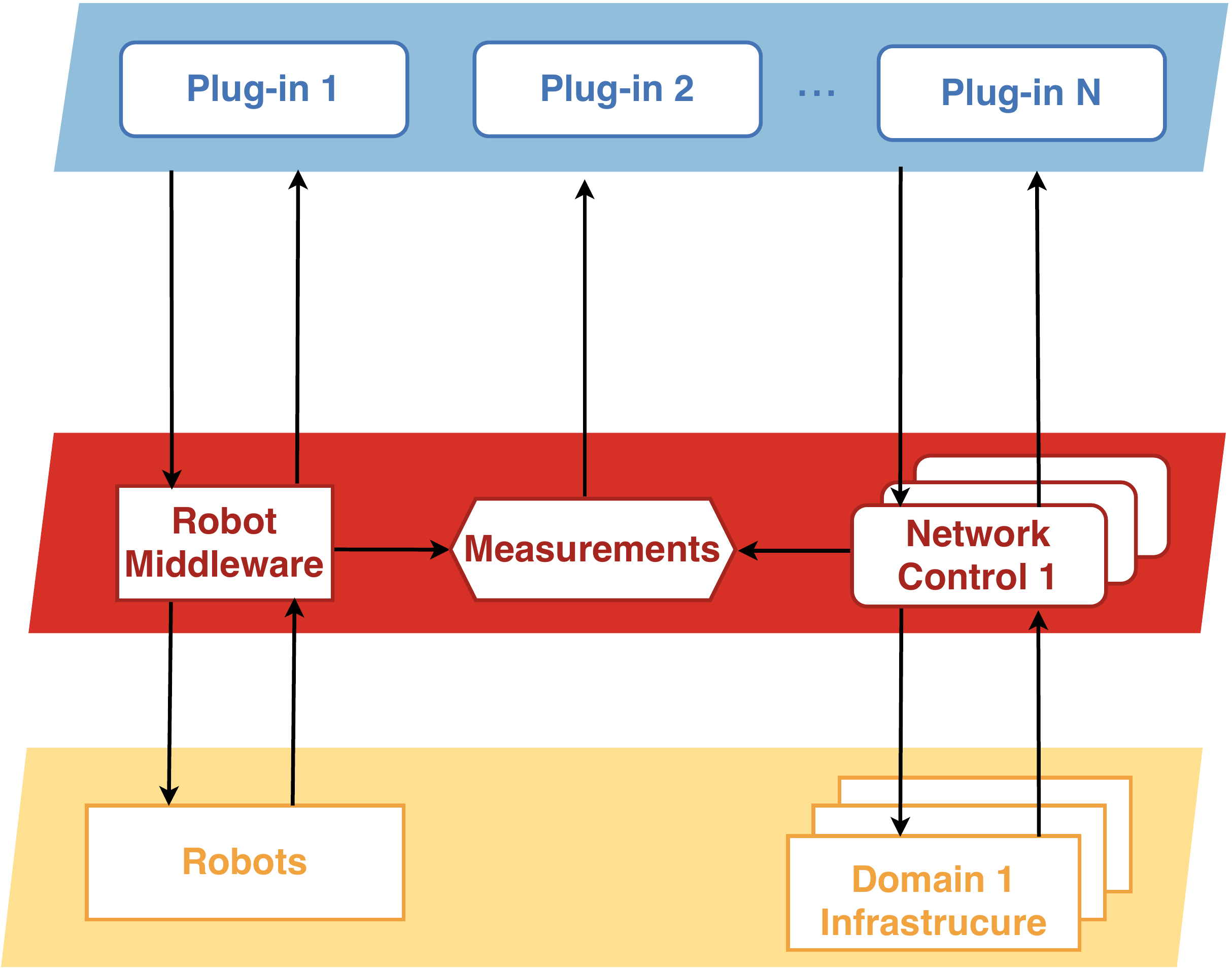}};
        % MATH SYMBOLS
        % 1. BELOW LAYER
        \node[fill=white,draw=gray] at (-3.6, \below) {$\phi(\cdot),\sigma_p(\cdot)$};
        \node[fill=white,draw=gray] at (-2, \below) {$\phi(\cdot),\sigma(\cdot)$};
        \node[align=right,fill=white,draw=gray] at (1.5, \below) {$a(\cdot),\delta(\cdot),\psi(\cdot)$};
        \node[align=right,fill=white,draw=gray] at (3.8, \below) { $\iota(\cdot), \lambda(\cdot), d(\cdot)$};
        % 1. ABOVE LAYER
        \node[fill=white,draw=gray] at (-3.6, \aboveBelow) {$\phi(\cdot),\sigma_p(\cdot)$};
        \node[fill=white] at (-2.3, \aboveBelow) {$\kappa(\cdot)$};
        % Middle layer interactions
        \node[fill=white,draw=gray,inner sep=1] at (-1.65, \middleAbove) {$\kappa(\cdot)$};
        \node[fill=white,draw=gray,inner sep=1,align=left] at (1.35, \middleAbove+.2) {$a(\cdot)$\\$d(\cdot)$\\$\lambda(\cdot)$};
        
        \node[align=right,fill=white,draw=gray] at (1.5, \aboveBelow) {$a(\cdot),\delta(\cdot),\psi(\cdot)$};
        
        \node[fill=white,draw=gray] at (0, \aboveAbove) {$\{\kappa^t(\cdot),a^t(\cdot),d^t(\cdot),\lambda^t(\cdot)\}_t$};
        \node[align=right,fill=white,draw=gray] at (3.8, \aboveBelow) { $\iota(\cdot), \lambda(\cdot), d(\cdot)$};
    \end{tikzpicture}
    \caption{COTORRA architecure}
    \label{fig:system}
\end{figure}
% link: https://app.diagrams.net/#G17BeSjZnvri7ikJm3rE_lGHDlMIgzvAkW

\todo[inline,disable]{Milan: edit the figure to: (i) show real time data consumption and historical data consumption, (ii) add link between the robot COTORRA mgmt and the measurements, (iii) point out the close loop feature in system. Both the robot closed loop and the infrastructure one}

% The system model of COTORRA is with modular design as shown on Fig.~\ref{fig:system}. The top (red) layer is designed to  attach external plug-ins to the COTORRA core layer (middle).
% \todo[inline]{MILAN: what about defining here also the plug-in layer that i represented as set of control and management algorithms ? Like that in the first paragraph we define the hardware layer and plug-ins layer and then we proceed with the COTORRA core where our main focus is
% }
% The core layer orchestrates and controls hardware (bottom layer) represented as a weighted graph $G$ with associated  embeddings holding its state.
% The hardware graph comprises robots $r_i\in G$, RUs $R_i\in G$,
% switching entities $w_i\in G$, and servers $s_i\in G$.

COTORRA system model follows a modular design as
shown in Fig.~\ref{fig:system}.
The hardware layer (bottom) is a weighted graph $G$
with graph embeddings\footnote{COTORRA uses
node $f(n),n\in V(G)$ and edge embeddings
$f(n_1,n_2), (n_1,n_2)\in E(G)$ of the hardware
graph $G$.}
that capture the state of
robots $r_i\in V(G)$
and hardware components present in a domain
infrastructure, i.e., Radio Units (RUs) $R_i\in V(G)$,
switching entities $w_i\in V(G)$, and servers $s_i\in V(G)$.
The plug-in layer is a set of user-defined
algorithms to control and manage both the robots and the infrastrucure.
The COTORRA core layer (middle) is a set of
building blocks enhancing the interaction
between the plug-in (top) layer and hardware (bottom) layer.

\subsection{COTORRA core layer building blocks}
\label{subsec:cotorra_core}

%% THIS IS THE INTRO
%% PARAGRAPH REMOVED
%In general, COTTORA serves as framework that enables closed-loop functionalities. The close loop starts when the hardware components send data to the core layer of COTTORA. Besides \st{gathering and }storing the data, the core layer of COTTORA implements mechanism for real-time data consumption \textcolor{red}{- this makes no sense}. The plug-ins algorithm analyze and combines the data to close the loop. They use the COTORRA core to send instructions to the robots or make modifications against  the underlying infrastructure. 
\todo[inline,disable]{KIRIL: After the first sentence, everything is confusing. Can we just say that the plugins are applications that can be used for robotic services e.g., that send instructions to the robots and receive feedback from the sensors via the COTORRA core. And to note that the core layer is crucial inter-working layer that makes it easier to extend the hardware or plug different plugins as well as emulate different conditions to test the dedicated plugins. And then explain the building blocks of the core layer that do so. Otherwise I get lost.

Milan: we remove the fist paragraph of this subsection
}

This section presents the functionality of each COTORRA
core layer building block, and explains the exchange
of information among layers and COTORRA core layer
building blocks.

\begin{itemize}
    \item The \textit{Robot Middleware}
    (\emph{a}) collects, (\emph{b}) stores and (\emph{c}) exposes real-time
    $\{r_i\}_i\subset V(G)$ robots' data;
    and (\emph{d}) sends instructions to the robots, e.g.,
    robots' movements.
    The \textit{Robot Middleware}
    periodically collects
    the robots' Radio Unit (RU) attachment
    $\phi(r_i, R_i)\in\{0,1\}$,
    and a vector of sensor data
    $\sigma(r_i)~\in~\mathbb{R}^n$ holding
    values such as 
    the robot position
    $\sigma_p(r_i),~2\le p\le d$, with $d$ being
    the space dimension.\footnote{Wheeled robots move
    on a $d=2$ space, and UAVs on a $d=3$ space.}
    Using the aforementioned information,
    the \textit{Robot Middleware} creates a contextual embedding~$\kappa(r_i)$
    \todo[inline,disable]{KIRIL: Maybe is me, but I don't understand the meaning of "elaborates a contextual embedding". Does it mean like gathers and stores in to a json-like structure? Or it is something else? It confuses me. Can it be: "creates a contextual structure"
    
    JORGE: an embedding is a function in the graph domain,
    e.g., in this case it is a function taking a node (robot)
    as input, and giving as output a vector. This function
    is the contextual embedding.
    }
    \begin{align}
        \kappa \colon \{r_i\}_i &\to \mathbb{R}^n\\
        r_i &\mapsto \big(\phi(r_i,R_1),
        \dots, \phi(r_i,R_{N}), \sigma(r_i)\big) \notag
    \end{align}
    %$\kappa: \{r_i\}_i \mapsto \mathbb{R}^n$
    %(e.g. $\kappa(r_1)~=~\left(\rho(r_1), \phi(r_1,R_1), \omega(r_i), \psi(r_i)\right)$)
    that is stored
    (via
    the \textit{Measurements} API)
    in
    the \textit{Measurements} building block,
    and exposed (via its own API) to the
    plug-in layer (see Fig.~\ref{fig:system}).
    Latter, a plug-in uses the contextual
    embedding to elaborate real-time 
    navigation $\sigma_p(r_i)$ (e.g., robot movement) and RU attachment 
    instructions $\phi(r_i,R_i)$, and forwards them to
    the \textit{Robot Middleware} API.
    Finally, \textit{Robot Middleware}
    sends the plug-in instructions
    to the robots.
    %Additionally, the
    %\textit{COTORRA robot mgmt}
    %module receives movement $\rho(\cdot)$ and
    %RU attachment instructions $\phi(\cdot)$ that
    %it latter sends to the robots.
    %{\color{red} Rewrite to emphatize that
    %only $\rho(),\phi()$ relate to decisions.}

    \item The \textit{Network Control}
    (\emph{a}) collects, (\emph{b}) exposes, and (\emph{c}) stores
    network and RU context information.
    One of its main functionalities is to (\emph{d}) emulate
    network conditions.
    As well, it (\emph{e}) maps Virtual Network Functions (VNF) and
    Virtual Links (VL) in the hardware graph,
    (\emph{f}) stores, and
    (\emph{g}) exposes the VNF/VL mappings.
    The \textit{Network Control} collects a vector with
    RUs' context information $\iota(R_i)\in\mathbb{R}^m,~R_i\in V(G)$,
    so as the links' throughput
    $\lambda(n_1,n_2),~(n_1,n_2)\in E(G)$, and delay
    $d(n_1,n_2),~(n_1,n_2)\in E(G)$ between every
    robot, RU, switch, and server
    in the hardware graph $G$.
    In addition, the \textit{Network Control} exposes 
    the real-time RU context information,
    links' delay, and throughput
    via its API; and stores
    the prior data
    in the \textit{Measurements} entity using the \textit{Measurements} API.
    To emulate network conditions, 
    \textit{Network Control} (\emph{i}) retrieves
    $\lambda(\cdot)$ throughput, and
    $d(\cdot)$ delay  embeddings from
    the \textit{Measurements} entity; and (\emph{ii})
    introduces artificial queuing
    delay
    $\psi(n_1,n_2)\ge0,~(n_1,n_2)\in E(G)$
    so as packet drop rate
    $\delta(n_1,n_2)\ge0$.
    As a result, the links
    $(n_1,n_2) \in E(G) $ in the hardware
    graph offer
    a delay of $\psi(n_1,n_2)d(n_1,n_2)$, and
    a throughput of
    $\frac{1}{\delta(n_1,n_2)}\lambda(n_1,n_2)$.
    The \textit{Network Control} API exposes
    operations to request both, queueing
    delay, and packet~drop.
    \vspace{0.5em} \\ %
    Additionally, the \textit{Network Control}
    acts as a Virtual Infrastructure
    Manager (VIM)~\cite{mec-nfv} that
    (\emph{i}) maps VNFs $\{v_i\}_i$ to hardware nodes
    $a(n)=\{v_1,\ldots,v_n\}, n\in V(G)$; and (\emph{ii}) 
    maps VLs in the hardware links
    $a(n_1,n_2)=\{(v_1,v_2),\ldots, (v_{n-1},v_n)\}$.
    Both VNFs, and VLs mappings are
    periodically (\emph{h}) stored
    and (\emph{i}) exposed by the
    \textit{Measurements} component
    (using the \textit{Measurements} API); and
    plug-ins can issue new mappings
    over the exposed
    \textit{Network Control} API.
    \item The \textit{Measurements} building block
    (\textit{a}) stores, and (\textit{b}) exposes a
    history of every embedding.
    The embeddings history is represented
    as a set 
    \begin{equation}
        \left\{ \{\kappa^t(r_i)\}_{r_i},
        \{a^t(n)\}_n,
        \{d^t(n_1,n_2)\}_{n_i},
        \{\lambda^t(n_1,n_2)\}_{n_i}
        \right\}_t
        \label{eq:historic}
    \end{equation}
    with $\kappa^t(r_i)$ denoting
    the contextual embedding of robot
    $r_i$ at time $t$. As shown in (\ref{eq:historic}), the
    embeddings history stores
    the VNF/VL mappings,
    network delay, and throughput information.
    The embeddings history is exposed via
    the \textit{Measurements} API
    to the plug-in layer (see
    Fig.~\ref{fig:system}).
\end{itemize}
%{\color{red} explain that remaining component handle the links
%embeddings, and see where to put measurements}

\subsection{COTORRA applicability: an example use case}
\label{subsec:applicability}
COTORRA is designed following the principles and definitions of some of the most known implementations of Edge and Fog computing such as Multi-access Edge Computing (MEC)~\cite{etsi.mec} and OpenFog working group in Industrial Internet Consortium (IIC)~\cite{icc}. COTORRA building blocks, namely, \textit{Robot Control}, \textit{Network Control}, \textit{Measurements} and \textit{Plug-ins} can be represented as autonomous applications and context-aware services running on top of a virtualization infrastructure. This implies that by using COTORRA we are contributing to the experimental evaluation of MEC and IIC compliant applications.

Lets consider an Edge robotic autonomous navigation scenario where the robot driver VNF $v_r$ runs on the robot $r_i$, and an autonomous navigation VNF $v_d$ runs on an Edge server $s_i$. A plug-in can use the \textit{Network Control} and \textit{Measurements} APIs to obtain RUs' context information $\iota(R_i)$ (e.g., RSSI, throughput, etc.) and \textit{Robot Middleware} to obtain robot sensor data $\sigma(r_i)$ (e.g., odomety , LIDAR, and camera data). Based on this information plug-ins can be designed, developed and tested with autonomous navigation algorithms that through the \textit{Robot Middleware} APIs can adapt the robot speed (present inside~$\sigma(r_i)$) based on the radio link status $\iota(R_i)$. Additionally, packet drop $\delta(n_1,n_2)$ and queuing
delay $\psi(n_1,n_2)$ can be emulated using the \textit{Network Control} to simulate congested network environments.

\section{COTORRA implementation}
\label{sec:implementation}
\begin{figure}
     \includegraphics[width=\columnwidth]{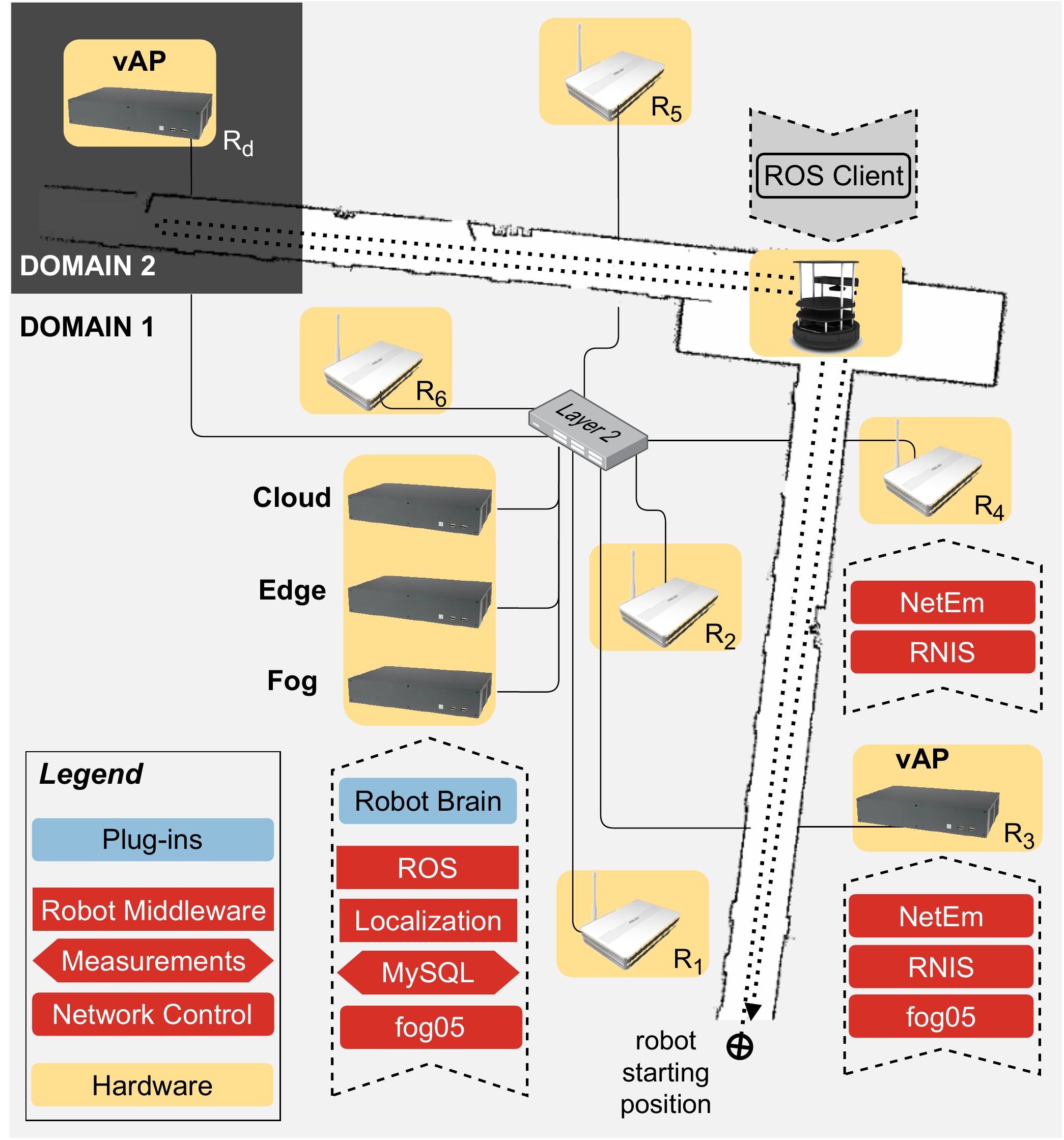}
    \caption{COTORRA testbed in Universidad Carlos
    III de Madrid facilities.}
    \label{fig:testbed}
\end{figure}
Fig.~\ref{fig:testbed} shows an implementation of COTORRA testbed based on a mobile robot. This implementation is used to verify the effectiveness of COTORRA in testing innovative mobility features. The COTORRA testbed consists  of: (\emph{i}) x5 ASUS  WL500G Premium v1 Access Points (APs) running OpenWrt 18.06.2; (\emph{ii}) x5 MiniPC, with x4 vCPUs and 8GB of RAM each. Two MiniPCs are used as APs and the rest as servers. Two of the server MiniPCs simulate Edge and Cloud nodes by introducing an artificial queuing delay of 5 and 10 seconds, respectively. The third MiniPC server acts as a local Fog node. All APs and MiniPCs are deployed along a corridor of the Universidad Carlos III de Madrid, interconnected with 10Gbps Ethernet connectivity. For the mobile robot, we used a ROS-compatible Kobuki Turtlebot~S2 robot equipped with a laptop with 8-GB RAM and 2 CPUs, and a RPLIDAR A2 lidar for 360-degree omnidirectional laser range scanning.

The COTORRA core layer is implemented using VNFs (shown as red rounded boxes on Fig.~\ref{fig:testbed}). The \textit{Robot Middleware} is implemented using Robotic Operating System (ROS) to (\emph{i}) collect and expose the lidar and odometry sensor data $\sigma(r_i)$ to the plug-ins layer, and (\emph{ii}) send to the robots the navigation instructions $\sigma_p(r_i)$ received from the plug-ins. In addition, as part of the \textit{Robot Middleware}, we use a Localization service to provide probabilistic robot localization
$\mathbb{P}\left(\sigma_p(r_i)\right)$ based on the lidar data. The \textit{Network Control} is implemented using a WLAN Access Information Service (WAIS) to provide real-time context information (e.g., signal level $\iota(R_i)$, transmission and reception bit rates $\lambda(r_i,R_i)$, etc.) through a REST API. In addition, \textit{Network Control} emulation is implemented with NetEm, so as to introduce artificial latency and packet loss.
And for the \textit{Network Control} VIM functionalities, we have used Fog05\footnote{Fog05 project: https://fog05.io/ [Accessed: 26 November 2020]} which implements the mapping
functionality mentioned in
Sec.~\ref{subsec:cotorra_core}. The \textit{Measurements} embeddings history~(\ref{eq:historic})
is stored and exposed with a a MySQL database.
Finally, our COTORRA implementation used
a ROS brain as a plug-in to provide autonomous navigation for the mobile robot through the \textit{Robot Middleware}.
\section{Experimental validation}
\label{sec:validation}
%In this section we show how COTORRA as a testbed can tackle arbitrary robotic scenarios. The implemented prototype of COTORRA from Sec.~\ref{sec:implementation} is used for step by step evaluation of orchestration and federation concepts for robotic services.
In this section we validate COTORRA testbed
by an evaluation of (an orchestration and
a federation) applications on top of the
COTORRA implementation described in
Section~\ref{sec:implementation}.

\subsection{Service orchestration that meets key performance indicators}
\label{subsec:okpi}
OKpi~\cite{okpi} is a service orchestration heuristic that
meets latency, and mobility requirements.
OKpi was implemented as a plug-in interacting with
COTORRA to orchestrate an autonomous navigation service
as described in Section~\ref{subsec:applicability}.
In the experiment, OKpi used the Localization service
to obtain the real-time robot $r$ position $\sigma_p(r)$,
and MySQL for network links delays' $d(n_1,n_2)$. Based on
that, it sent
(i) to fog05 the autonomous navigation VNF mapping
\mbox{$a(s_i)=\{v_d\}, s_i\in\{\text{Fog, Edge, Cloud}\}$},
and (ii)
% the APs attachment $\phi(r,R_i)$
802.11r handovers $\phi(r,R_i)$
to the robot $r$
as it traveled along a corridor
(see Fig.~\ref{fig:testbed} dashed trajectory).

\begin{figure}[t]
    \begin{tikzpicture}[font=\tiny]
        % PLOT
        %%% original version
        %\node[inner sep=0pt] (arch) at (0,0) {\includegraphics[width=\columnwidth]{img/okpi-results.eps}};
        %%% arXiv version -> no EPS supported
        \node[inner sep=0pt] (arch) at (0,0) {\includegraphics[width=\columnwidth]{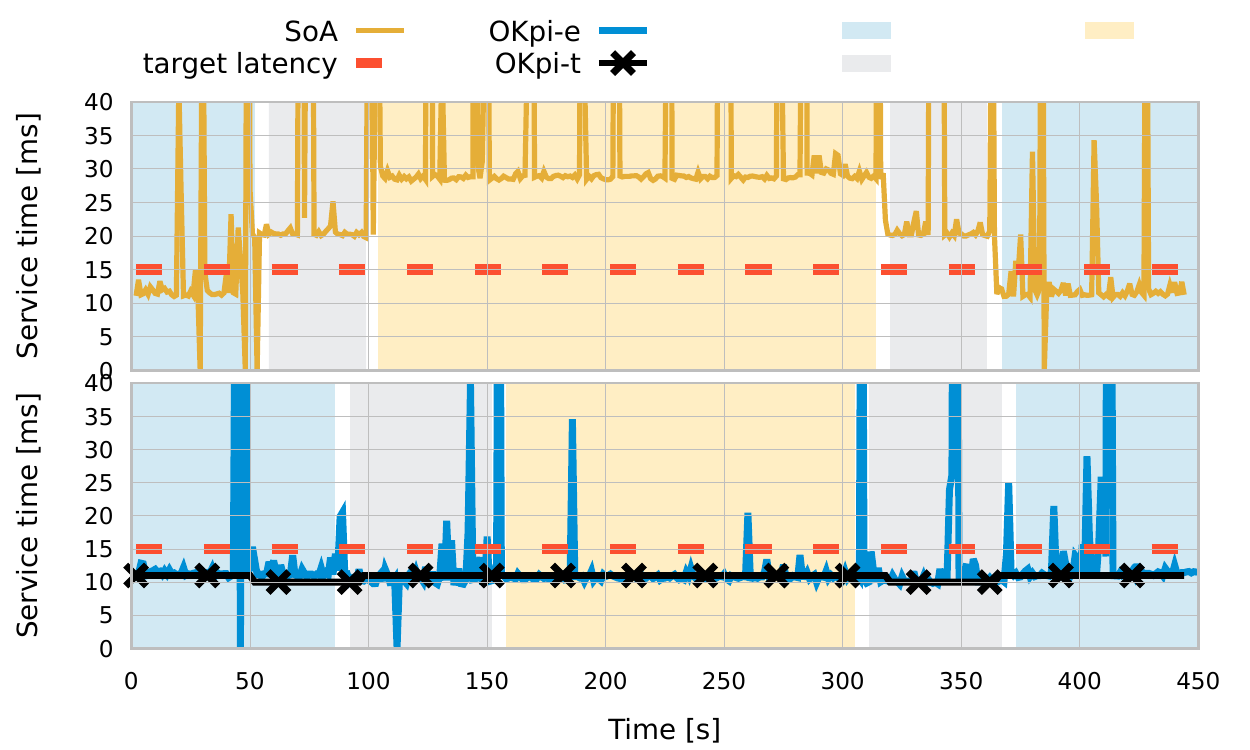}};
        \node at (.9, 2.4) {$\phi(r,R_{1,3})$};
        \node at (.9, 2.2) {$\phi(r,R_{2,4})$};
        \node at (2.65, 2.4) {$\phi(r,R_{5,6})$};
    \end{tikzpicture}
    \caption{Using COTORRA to compare a SoA and OKpi autonomous navigation.}
    \label{fig:okpi}
\end{figure}
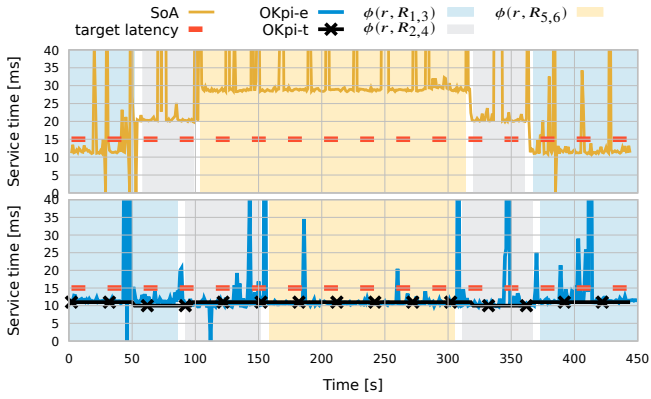

The experiment goal was to compare
a State Of the Art (SoA) autonomous driving with
% 802.11r <- before
periodic probing for RU attachment,
against the theoretical
\mbox{\textit{OKpi-t}} and empirical \mbox{\textit{OKpi-e}} performance
of OKpi; and if the latter achieved a service time\footnote{%
Elapsed time between the emission of
robot position packets, and the
reception of robot navigation packets.
}
below 15~ms.
Fig.~\ref{fig:okpi} illustrates (at the top) that
the service time remained most of the time
over 15~ms when using
% SoA 802.11r RU attachment, 
the SoA autonomous driving,
and $a(\text{Cloud})=\{v_d\}$;
whilst at the bottom, both the theoretical, and empirical
performance of OKpi remained below 15~ms except for
the handover peaks. Additionally, Fig.~\ref{fig:okpi}
shows the RUs attachments over time. 

\subsection{DLT federation}

% In~\cite{antevski2020dlt}, COTORRA is used to showcase a federation of a robotic service using a Distributed Ledger Technology (e.g., Blockchain). 
In~\cite{antevski2020dlt}, we designed a federation of a robotic service using a Distributed Ledger Technology (e.g., Blockchain). 
Service federation is a set of procedures that enable orchestration of services across multiple administrative domains. 
We used COTORRA testbed to deploy the DLT solution and perform experimental evaluation of the multi-domain federation. 
The experiment consisted of a moving robot towards out-of-coverage area of Domain 1 (on Fig.~\ref{fig:testbed}) and extending the connectivity range, on-the-fly, in  Domain 2. 
An \emph{orchestrator} plug-in was deployed in each domain, and a DLT plug-in was used for the inter-domain interaction.

\begin{figure}[t]
    \begin{tikzpicture}[font=\tiny]
        \pgfmathsetmacro{\rightDLT}{3.3}
        \pgfmathsetmacro{\leftDLT}{-1.5}
        \pgfmathsetmacro{\aboveBelow}{0.8}
        % PLOT
        \node[inner sep=0pt] (arch) at (0,0)
        % PLOT
        {\includegraphics[width=\columnwidth]{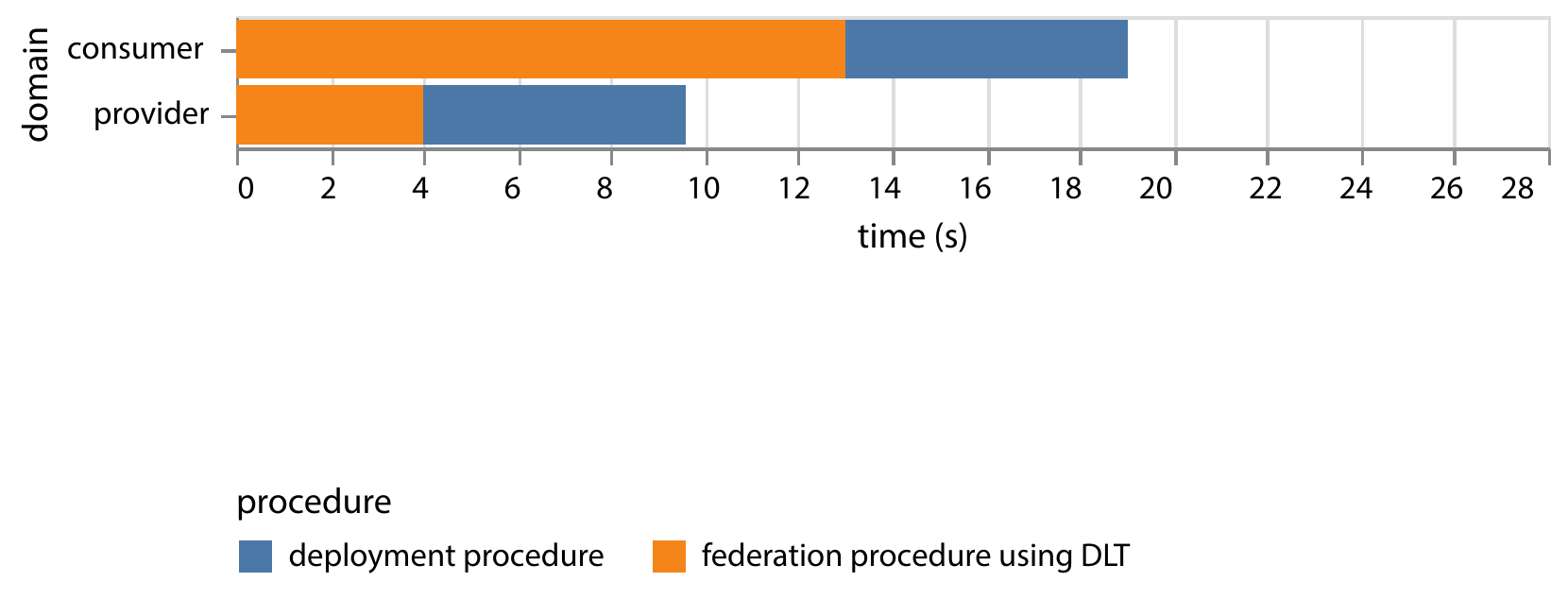}};
        \draw[thick, ->] (\leftDLT-1.55,0.48) arc (-180:-90:0.5);
        \node[fill=white,draw=gray,text width=1.8cm,align=left] at (\leftDLT, -0.3) {$a(R_3)=\{vAP\}$\\$a(R_d)=\{\}$\\\ $\phi(r)=R_3$};
        \draw[thick, ->] (\rightDLT-1.35,0.48) arc (-180:-110:0.5);
        % {$\{\kappa^t(\cdot),a^t(\cdot),d^t(\cdot),\lambda^t(\cdot)\}_t$};
        \node[fill=white,draw=gray,text width=1.8cm,align=left] at (\rightDLT, -0.3) {$a(R_3)=\{\}$ \\$a(R_d)=\{vAP\}$ \\ $\phi(r)=R_d$};
    \end{tikzpicture}
    \caption{Using COTORRA to measure DLT-negotiation
    times, and vAPs deployment times in a federated
    environment.}
    \vspace{-5mm}
    \label{fig:dlt_results}
\end{figure}
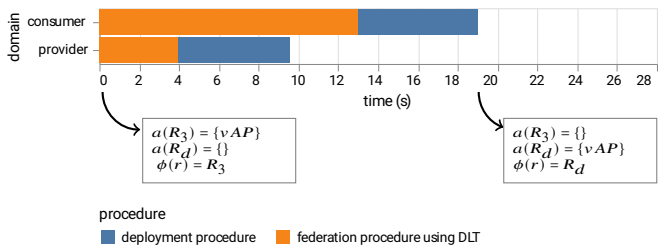

The mobile robot moved from a starting position (similar as on Fig.~\ref{fig:testbed}) towards the coverage area of Domain 2. The Robot brain, analyzed the location data from the $\sigma_p(r)$ Localization service, and triggered a federation procedure through the Domain 1 orchestrator. Both orchestrators negotiated and established deployment
of a vAP through the DLT plug-in, and the deployment decision
$a(n_{AP})=\{v_{AP}\}$ was sent to fog05.
The mobile robot attached $\phi(r,n_{AP})=1$ to the newly deployed vAP in Domain 2 without any movement or robotic service interruption. As shown on Fig.~\ref{fig:dlt_results}, it took around 19 seconds to complete the described federation procedure.
% A DLT/Blockchain plug-in and an \emph{orchestrator} plug-ins  are deployed in each domain to perform the federation process over . 
% , the combined information from Localization enabled the Robot brain plug-in to trigger a federation procedure to an Orchestrator, additional plug-in that orchestrates the robotic service in Domain 1. 

% How to include the multi domain into COTTORA?
% In the following experiment,

% 1-paragraph to explain the results

\section{Conclusion}
\label{sec:conclusion}
In this letter, we introduce COTORRA, a modular tested built to enable and 
% demonstrate the benefits of the proposed COTORRA testbed 
validate innovative mechanisms or applications in robotic system. 
Additionally, we showed that COTTORA can be easily implemented over commodity hardware that offers 
% experimental 
support for a rapid prototyping and validation. 
The network emulation feature in COTORRA can simulate unpredictable network conditions that enables near production robotic environment where new pluggable applications can be tested. 
% , which then results in more efficient robotic application validation.
% Experimental results show 
We validated this by running experiments for 
% that COTORRA can be used to test 
orchestration algorithms that interact with both the network infrastructure, and mobile robots. 
Finally, the flexibly and scalability of COTORRA enabled us to 
% Furthermore, COTORRA was used to 
showcase an application of DLT for federation in an Edge robotic service, guaranteeing service footprint extension.

%This letter presents COTORRA, a context-aware robotic testbed, and implements the proposed system design to verify its effectiveness. 
%Experimental results show that COTORRA can be used to test orchestration algorithms that interact with both the network infrastructure, and mobile robots. Furthermore, COTORRA was used to showcase an application of DLT for federation in an Edge robotic service, guaranteeing service footprint extension. 
% The contribution of this letter 
%Additionally, it describes a COTORRA implementation that offers experimental support for rapid prototyping robotic applications in Edge and Fog environment.    
%I met this girl, when I was ten years old.
%And what I loved most she had so much soul.
%She was old school, when I was just a shorty.
%Never knew throughout my life she would be there for me.
%On the regular, not a church girl she was secular.
%Not about the money, those studs was mic checking her.
%But I respected her, she hit me in the heart.
%A few New York niggas, had did her in the park.
%%But she was there for me, and I was there for her.
%Pull out a chair for her, turn on the air for her.
%Boy I tell ya, I miss her.
%But I'mma take her back hoping that the shit stop.
%Cause who I'm talking bout y'all is \textbf{hip-hop}.

% \section*{Acknowledgment}
% This work has been partially funded by the EU H2020
% 5GROWTH Project (grant no. 856709) and by the H2020
% collaborative Europe/Taiwan research project 5G-DIVE
% (grant no. 859881).

\ifCLASSOPTIONcaptionsoff
  \newpage
\fi

\bibliographystyle{IEEEtran}
\bibliography{main}

\end{document}